\documentclass[10pt,twocolumn,letterpaper]{article}

\usepackage[pagenumbers]{cvpr} %

\usepackage{graphicx}
\usepackage{amsmath}
\usepackage{amssymb}
\usepackage{booktabs,eucal}

\usepackage{caption}
\usepackage[dvipsnames,table,xcdraw]{xcolor}
\usepackage[pagebackref=false,breaklinks,colorlinks,citecolor=RoyalBlue,bookmarks=false]{hyperref}

\usepackage{nicefrac}       %
\usepackage{microtype}      %

\usepackage{blindtext}
\usepackage{float}
\usepackage{enumitem}
\usepackage[table]{xcolor}
\usepackage{tabulary,multirow,overpic}
\usepackage{verbatim}
\usepackage{color}
\usepackage{dblfloatfix}
\usepackage{pifont}%
\usepackage[accsupp]{axessibility}  %
\usepackage{makecell}

\usepackage{marvosym}

\usepackage{dblfloatfix}
\usepackage{nicefrac}

\usepackage{relsize}

\usepackage{listings}
\definecolor{linkcolor}{RGB}{255,0,0}
\definecolor{urlcolor}{RGB}{255,105,180}
\definecolor{citecolor}{RGB}{66,168,235}
\definecolor{codegreen}{rgb}{0,0.5,0}
\definecolor{codeblue}{rgb}{0.25,0.5,0.5}
\definecolor{codegray}{rgb}{0.6,0.6,0.6}

\newlength\savewidth\newcommand\shline{\noalign{\global\savewidth\arrayrulewidth
		\global\arrayrulewidth 1pt}\hline\noalign{\global\arrayrulewidth\savewidth}}
\newcommand{\tablestyle}[2]{\setlength{\tabcolsep}{#1}\renewcommand{\arraystretch}{#2}\centering\small}

\newcolumntype{x}[1]{>{\centering\arraybackslash}p{#1pt}}
\newcolumntype{y}[1]{>{\raggedright\arraybackslash}p{#1pt}}
\newcolumntype{z}[1]{>{\raggedleft\arraybackslash}p{#1pt}}

\definecolor{baselinecolor}{gray}{.92}

\definecolor{demphcolor}{gray}{.2}

\definecolor{demphcolorinline}{gray}{.3}

\definecolor{demphcolor1}{gray}{.6}
\newcommand{\demphs}[1]{\textcolor{demphcolor1}{#1}}

\renewcommand{\paragraph}[1]{\vspace{1.25mm}\noindent\textbf{#1}}

\newcommand{\app}{\raise.17ex\hbox{$\scriptstyle\sim$}}

\newcommand{\cmark}{\ding{51}}%
\newcommand{\xmark}{\ding{55}}%
\newcommand{\authorskip}{\hspace{5mm}}

\usepackage{marvosym}

\usepackage[capitalize]{cleveref}
\crefname{section}{Sec.}{Secs.}
\Crefname{section}{Section}{Sections}
\Crefname{table}{Table}{Tables}
\crefname{table}{Table}{Tables}

\def\confName{CVPR}
\def\confYear{2023}

\def\Ours{{Painter}\xspace}

\usepackage[accsupp]{axessibility}  %

\begin{document}

\title{Images Speak 
in Images: A Generalist Painter for In-Context Visual Learning}

\author{Xinlong Wang\textsuperscript{1}\thanks{Equal contribution. Correspondence to \textit{xinlong.wang96@gmail.com}. This work is done when Wen Wang is an intern at BAAI. } 
\authorskip Wen Wang\textsuperscript{2}$^*$ \authorskip Yue Cao\textsuperscript{1}$^*$ \authorskip 
Chunhua Shen\textsuperscript{2} \authorskip Tiejun Huang\textsuperscript{1,3} \\[2mm]
{
\fontsize{10.4pt}{9.84pt}\selectfont
\textsuperscript{1} Beijing Academy of Artificial Intelligence \hspace{5.7mm} 
\textsuperscript{2} Zhejiang University \hspace{5.7mm} \textsuperscript{3} Peking University}\\[1.5mm]
{
\fontsize{9.4pt}{9.84pt}\selectfont
\url{https://github.com/baaivision/Painter}
}
}

\input figs/fig1.tex

\maketitle

\begin{abstract}
In-context learning, as a new paradigm in NLP, allows the model to rapidly adapt to various tasks with only a handful of prompts and examples. But in computer vision, the difficulties for in-context learning lie in that tasks vary significantly in the output representations, thus it is unclear how to define the general-purpose task prompts that the vision model can understand and transfer to out-of-domain tasks. 
In this work, we present \Ours, a generalist model which addresses these obstacles with an ``image''-centric solution, that is, to redefine the output of core vision tasks as images, and specify task prompts as also images.
With this idea, our training process is extremely simple, which performs standard masked image modeling on the stitch of input and output image pairs. This makes the model capable of performing tasks conditioned on visible image patches. Thus, during inference, we can adopt a pair of input and output images from the same task as the input condition, to indicate which task to perform.
Without bells and whistles, our generalist \Ours can achieve competitive performance compared to well-established task-specific models, on seven representative vision tasks ranging from high-level visual understanding to low-level image processing. 
In addition, \Ours significantly outperforms recent generalist models on several challenging tasks.

\end{abstract}

\section{Introduction}
\label{sec:intro}
Training one generalist model that can 
execute 
diverse tasks simultaneously, and even can perform a new task given a prompt and very few examples, is 
one important step closer to artificial general intelligence. 
In NLP, the emergence of in-context learning~\cite{gpt3,Alayrac2022flamingo,Hao2022metalm} presents a new path in this direction, which uses language sequences as the general interface and allows the model to rapidly adapt to various language-centric tasks with only a handful of prompts and examples.

Thus far, 
in computer vision, in-context learning is 
rarely 
explored and remains %
unclear how to achieve that. 
This can be %
attributed to 
the differences between two modalities. One difference is that NLP tasks mainly consist of language understanding and generation, so their output spaces can be unified as sequences of discrete language tokens. But vision tasks are the abstractions of raw visual input to varied granularity and angles. Thus, \emph{vision tasks vary significantly in output %
representations}, leading to %
various task-specific 
loss functions and architecture designs. %
The second difference is that the output space of NLP tasks is even the same as the input. Thus, the task instruction and the example's input/output, which are all language-token sequences, can be directly used as the input condition (also denoted as the task prompt), which %
can be processed straightforwardly 
by the large language model. %
However, in computer vision, it is unclear how to \emph{define general-purpose task prompts or instructions} that the vision model can understand and transfer to out-of-domain tasks.
Several recent %
attempts 
\cite{Chen2021pix2seq,Chen2022pix2seq2,Kolesnikov2022UVIM,Lu2022unifiedio,Wang2022OFA,Bar2022VisualPrompt} %
tackle 
these difficulties by following %
the 
solutions in NLP.
They more-or-less convert vision problems into NLP ones via
discretizing the continuous output spaces of vision tasks, and using the language or specially-designed discrete tokens as the task prompts.

However, we believe that images speak in images, \ie, image itself is a natural interface for general-purpose visual perception.
In this work, we address the above obstacles with a vision-centric solution. %
The core observation is that most dense-prediction vision 
problems can be formulated as \textit{image inpainting}, \textit{i.e.}:

{\it 
Given an input image, prediction is to inpaint the desired but missing output ``image''.
}

Thus, we need a representation 
 of 3-channel tensor that appears as an ``image''   
for the output of the vision tasks, and specify the task prompts 
using a pair of images. 
Here we showcase several representative vision tasks for training, including depth estimation, human keypoint detection, semantic segmentation, instance segmentation, image denoising, image deraining, and image enhancement, and unify their output spaces 
using a 3-channel tensor, \textit{a.k.a.}\  ``output image''. 
We carefully design the data format for each task, such as instance mask and per-pixel discrete labels of panoptic segmentation, per-pixel continuous values of depth estimation, and high-precision coordinates of pose estimation.
Including more tasks is very straightforward, as we 
only need to construct new data pairs and add them to the training set, without modifications
to either the model architecture or loss function.

Based on this unification, we train a generalist \Ours model with an extremely simple training process.
During training, we stitch  
two images from the same task into a larger image,  and so do their corresponding output images.
Then we apply masked image modeling (MIM) on pixels 
of the output image, with the input image being the condition.
With such a learning process, 
we enable the model to perform tasks conditioned on visible image patches, that is, the capability of in-context prediction with the visual signal as context.

Thus the trained model is capable of the in-context inference. That is, we directly use the input/output paired images from the same task as the input condition to indicate which task to perform. Examples of in-context inference are illustrated in Figure~\ref{fig:teaser}, consisting of seven in-domain examples (seven rows at top) and three out-of-domain examples (three rows at bottom).
This definition of task prompts does not require deep understanding of language instructions 
as need by almost all previous approaches, 
and makes it very flexible for performing both in-domain and out-of-domain vision tasks.

Without bells and whistles, our model can achieve competitive performance compared to well-established task-specific models, on several 
fundamental 
vision tasks across high-level visual understanding to low-level image processing, namely, depth estimation on NYUv2~\cite{nathan2012nyuv2}, semantic segmentation on ADE-20K~\cite{zhou2018ade}, human keypoint detection on COCO~\cite{lin2014coco}, panoptic segmentation on COCO~\cite{lin2014coco}, and three low-level image restoration tasks. Notably, on depth estimation of NYUv2, our model achieves state-of-the-art performance, outperforming previous best results 
by large margins which have heavy and specialized designs on architectures and loss functions. 
Compared to other generalist models, \Ours yields significant improvements on several challenging tasks.

\section{Related Work}

\paragraph{Unified Modeling}~~~The emergence of Transformer~\cite{vaswani2017attention} provides the possibility to share the basic modeling module across different modalities. Until now, Transformers are widely-adopted in language~\cite{devlin2018bert,liu2019roberta,gpt3}, vision~\cite{dosovitskiy2020vit,liu2021swin,carion2020detr,Chen2021pix2seq,bachmann2022multimae}, speech~\cite{dong2018speech,gulati2020conformer,karita2019comparative} and multimodal~\cite{lu2019vilbert,su2019vlbert,kamath2021mdetr} domains.
Perceiver~\cite{jaegle2021perceiver} and Perceiver-IO~\cite{Jaegle2021perceiverio} are the first attempts to use the exact same Transformer architecture in different domains, such as natural language and visual understanding processing, StarCraft II, and multi-modal domains. If the input could be transformed to a sequence of tokens, one can adopt Transformer for modeling the relationships between different tokens.

\paragraph{Vision Generalist}~~~Due to the general modeling capability of Transformer, there are some efforts to unify different tasks in vision domains, resulting in several vision generalists~\cite{Chen2021pix2seq,Chen2022pix2seq2,Zhu2022uniperceiver,Lu2022unifiedio,Kolesnikov2022UVIM}.
DETR~\cite{carion2020detr} first adopted Transformer as the task specific head for object detection. Based on this, Pix2Seq~\cite{Chen2021pix2seq} defined the output space of object detection as a discrete space, and conduct this task in an auto-regressive manner. Due to the fundamental nature of object detection, Pix2Seq provides a direction for unifying different vision tasks using discrete spaces, thus motivating a lot of following work.
Unified-IO~\cite{Lu2022unifiedio} and OFA~\cite{Wang2022OFA} both homogenize the diverse inputs and outputs to a sequence of discrete tokens, perform joint modeling in a sequence-to-sequence manner over vision, vision \& language and NLP tasks, and use T5-style architectures~\cite{t5} with billions of parameters, where Unified-IO unifies more tasks than OFA with larger size of models.
Pix2Seq v2 \cite{Chen2022pix2seq2} unified object detection, instance segmentation, keypoint estimation and image captioning in the same defined discrete spaces as Pix2Seq. 
UViM~\cite{Kolesnikov2022UVIM} unified pixel-labeling tasks with the same modeling approach but trained separate models for different tasks, such as panoptic segmentation, depth estimation and colorization.

Notably, from our point of view, the input of visual signals is continuous in nature, thus we try to make the output space of several representative vision tasks as continuous as images to reduce the quantization error caused by discretization and further enable the in-context visual learning with masked image modeling.

\paragraph{In-Context Learning}~~~For the first time, GPT-3~\cite{gpt3} defined a new learning paradigm, in-context learning, 
where 
a series of NLP %
tasks  
can be formulated as the text completion task given prompts and examples. In-context learning grants models new capabilities to perform on-the-fly computational reasoning or novel-pattern recognition that is unlikely to have occurred in training. Flamingo~\cite{Alayrac2022flamingo} extended the input of the large language models to not only texts but also images and videos, but still used languages as the general interface, such that the model can perform 
many 
visual-linguistic tasks given prompts and examples, such as image captioning, visual question answering, optical character recognition (OCR), \textit{etc}. 
In other domains, it 
appears 
non-trivial to directly introduce the in-context learning capability.
AD~\cite{Laskin2022promptRL} uses  algorithm distillation to combine in-context capability with reinforcement learning.
In computer vision, a concurrent work~\cite{Bar2022VisualPrompt} performs inpainting on the figures and infographics from vision articles, but only works for predicting on discrete space as the language domain, and shows the in-context capability on foreground segmentation, \textit{single  } object detection and colorization.
While the work \cite{Bar2022VisualPrompt} proves the concept 
of in-context learning for vision tasks, no results were reported on standard benchmark datasets. Thus, it remains unclear how the method performs on real-world datasets. 
In contrast, our model works well with masked image modeling on pixels on seven diverse and %
challenging 
vision tasks, including depth estimation, keypoint estimation, semantic segmentation, panoptic segmentation, image denoising, image deraining, and image enhancement, and also shows \textit{highly competitive} performance on these tasks.

\section{Approach}

The core idea of our framework is to reformulate most vision tasks such as depth estimation, semantic segmentation, instance segmentation, keypoint detection and image restoration as an image inpainting problem. To do so, we redefine the output space of those tasks as ``images''.

\subsection{Redefining Output Spaces as ``Images''}\label{sec:sub-redefine}

We denote an input image as $\bf{x}$ with the size of $H\times W\times 3$, and standard definition of the task ground truth as ${\bf y}^t$ which has various sizes for different task $t$, and we redefine these task outputs still in the image space, denoting as ${\hat {\bf y}}^t$ with the size of ${H}\times {W}\times 3$. Our philosophy is to keep the spatial relationships between pixels to be intact, and each pixel of the output image still represents the output for this task of the corresponding input image pixel but in the RGB space. That is, ${\hat {\bf y}}^t_{i,j}$ with three dimensions denotes the corresponding ground truth of input pixel ${\bf x}_{i,j}$. We select seven representative vision tasks with diverse types of outputs, such as depth estimation, semantic segmentation, keypoint detection and panoptic segmentation. Here we show how we redefine the per pixel ground-truth for each task as a 3-channel tensor, similar to %
the RGB space. Note that in theory, a fixed  number of output channels can serve our purpose. We choose 3 channels to make an image.

\paragraph{Monocular depth estimation} is a dense prediction task with weak semantics, to estimate the per-pixel depth value (distance relative to the camera) given an input RGB image. For NYUv2~\cite{nathan2012nyuv2}, the per-pixel ground-truth depth ${\bf y}^t_{i,j}$ is a real value in the range of $[0, 10]$ meters. Here we map the ground-truth value from real-valued range $ [0, 10] $ to the integer space with range $ [0, 255] $, ${\hat {\bf y}}^t_{i,j,0}=\lfloor{\bf y}^t_{i,j} \times \frac{255}{10}\rfloor$, 
and let the three channels ${\hat {\bf y}}^t_{i,j,0}$,  ${\hat {\bf y}}^t_{i,j,1}$ and  ${\hat {\bf y}}^t_{i,j,2}$ be the same ground truth. In inference, we directly average the outputs of the three channels and then perform the inverse linear transformation of the training to obtain a depth estimate in the range of $[0, 10]$.

\paragraph{Semantic segmentation} is a dense prediction task with strong semantics, to predict the per-pixel semantic label given an input image. Given a semantic segmentation task with $L$ categories, we let the RGB space to represent these $L$ categories with the same margin in each space. We represent $L$ as a 3-digit number with $b$-base system, where $b={\lceil L^{\frac{1}{3}}\rceil}$, and ${\hat {\bf y}}^t_{i,j,0}$, ${\hat {\bf y}}^t_{i,j,1}$ and ${\hat {\bf y}}^t_{i,j,2}$ represent their hundreds, tens and ones places, with a margin defined as $m=\lfloor\frac{256}{b}\rfloor$.
For example, ADE-20K~\cite{zhou2018ade} has 150 semantic categories with one background class, thus we set the base as $b=6$, and the margin as $m=42$. The output channels are defined as ${\hat {\bf y}}^t_{i,j,0}= \lfloor\frac{l}{b^2}\rfloor \times m$, ${\hat {\bf y}}^t_{i,j,1}=\lfloor\frac{l}{b}\rfloor \mod b \times m$, and ${\hat {\bf y}}^t_{i,j,2}= l \mod b \times m $, where $l$ denotes the corresponding category and is an integer value in range of $[0,L)$.
In inference, we discretize the output of each pixel with the margin $m$ and obtain its corresponding category.

\paragraph{Keypoint detection} is a fine-grained localization task to simultaneously detect the objects (\eg, human) and localize their detailed keypoints. We follow recent heatmap-based top-down pipeline~\cite{xiao2018sbaseline}, thus it is defined as a 17-category point-localization task for human keypoint detection. For each keypoint, we need to localize it to its corresponding pixel, which is very fine-grained. 
We decouple this task into a combination of 17-category keypoint classification using two channels of ${\hat {\bf y}}^t_{i,j,1}$ and ${\hat {\bf y}}^t_{i,j,2}$, and class-agnostic keypoint localization using another channel of ${\hat {\bf y}}^t_{i,j,0}$. 
For the 17-category classification task, we define each keypoint as a 9$\times$9 pixel square, and define the color of each square using the approach in semantic segmentation. 
For the class-agnostic point localization task, we define 17 squares, each of which is a 17$\times$17 heatmap with Gaussian distribution that the center pixel with largest value of 255 is the position of the ground truth keypoint. 
In inference, we obtain the category and location for each keypoint as the final results from the 3-channel output image.

\paragraph{Panoptic segmentation} is a combination of semantic segmentation task and instance segmentation task. Thus, we perform these two tasks separately for ease of redefinition of output space and optimization, and then combine their results to obtain the results for panoptic segmentation.

Here we introduce the redefinition of output space of class-agnostic instance segmentation. For this task, we directly change the color of each instance mask in the image to the same one, thus different instances use different colors. 
In theory, we can randomly choose colors for different instances, but we find that this setting would make the model hard to optimize. To address this optimization issue, we follow SOLO~\cite{wang2020solo} to assign the color of each instance mask according to the absolute position of its center in the image. 
We conceptually divide the image into
16$\times$20$\times$20 blocks, corresponding to three channels respectively.
We assign a fixed color to each block, then color the  mask accordingly if its center locates in that block.
In inference, we adopt each color as a kernel to compute the distance with each pixel in the image, and then set a threshold to get the final masks. To get the category for each instance mask, we directly assign the majority category in the semantic segmentation result within each instance mask as its category.

\paragraph{Image restoration} takes the corrupted image as input and outputs the corresponding clean image. 
In this paper, we investigate three representative image restoration tasks, including image denoising, image deraining, and low-light image enhancement.
Since both the input and output are inherently defined in the RGB space, these tasks can be seamlessly unified in our \Ours model without any transformation.

\begin{figure}
    \centering
    \includegraphics[width=.98\linewidth]{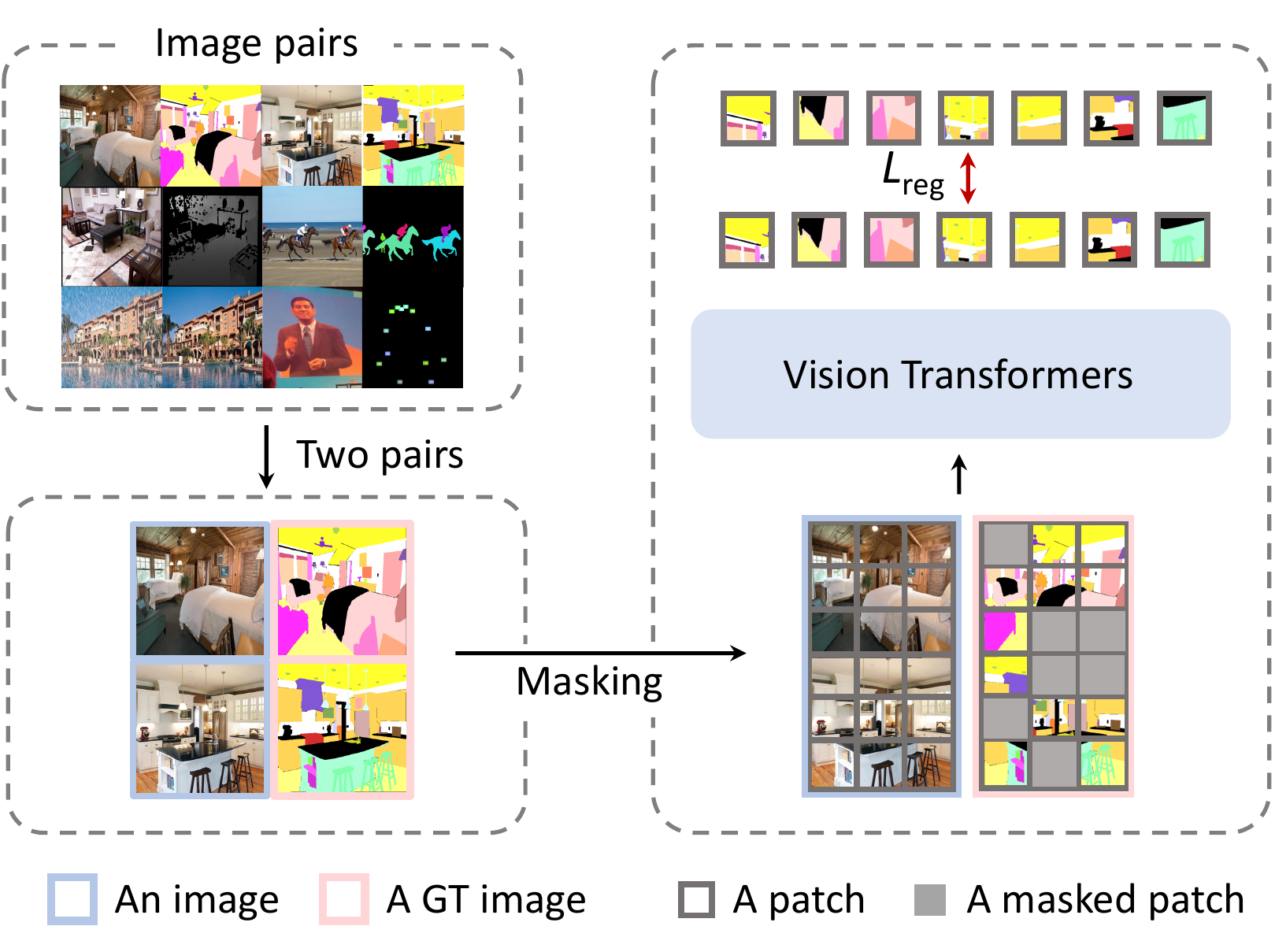}
    \vspace{-0.8em}
    \caption{
    {\bf The training pipeline of the masked image modeling (MIM) framework.
    } 
    }
    \label{fig:arch}
    \vspace{-1em}
\end{figure}

\subsection{A Masked Image Modeling Framework}\label{sec:sub-mim}

Based on the redefinition of the output spaces of above representative vision tasks, the input and output of these tasks are all images. Thus, here we directly apply a standard masked image modeling (MIM) pipeline~\cite{bao2021beit,xie2021simmim,he2021masked} for training, illustrated in Figure~\ref{fig:arch}. This framework consists of three major components: input format, architecture and loss function.

\paragraph{Input Format}~~~During training, each input sample is the concatenation of two pairs of images from the same task, which have already been applied data augmentations separately, as shown in the left part of Figure~\ref{fig:arch}. Each image pair consists of one image and its corresponding task output, which is also redefined as an image.
We randomly mask the task output image and train the model to reconstruct the missing pixels.
For the masked area, we follow the
NLP community~\cite{su2019vlbert,liu2019roberta} and previous works~\cite{bao2021beit,xie2021simmim,he2021masked} to use a learnable token vector to replace each masked patch. 
We adopt the block-wise masking strategy and find the masking ratio as 75\% to work well.

\begin{figure*}
    \centering
    \includegraphics[width=.85\linewidth]{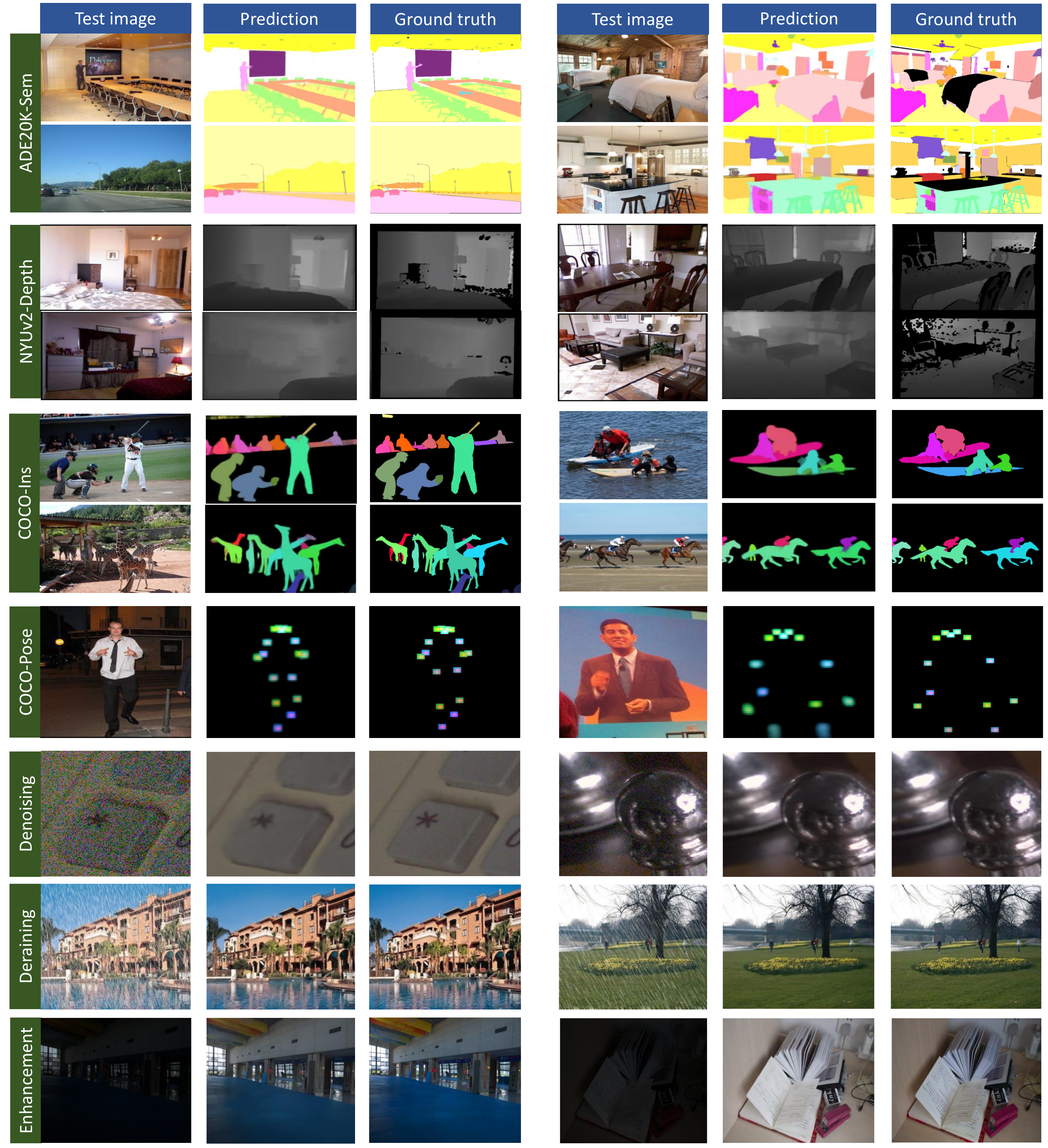}
    \vspace{-.5em}
    \caption{
    {\bf Visualizations of examples and predictions obtained by our \Ours for different tasks},
    such as 
    semantic segmentation, 
    depth estimation, %
    instance segmentation, %
    human keypoint detection, image denoising, image deraining and low-light image enhancement. Best viewed on screen.}
    \label{fig:vis}
    \vspace{-1.5em}
\end{figure*}

\paragraph{Architecture}~~~We adopt a vanilla vision Transformer (ViT)~\cite{dosovitskiy2020vit} as the encoder, which consists of stacked Transformer blocks, \eg, 24 blocks for ViT-large.
We concatenate 4 feature maps evenly sampled from these blocks and use a simple three-layer head to map the features of each patch to its original resolution, \eg, $16\times 16\times 3$.
Specifically, the head consists of a linear (1$\times$1 convolution) layer, a 3$\times$3 convolution layer, and another linear layer.

Since each input sample includes both the input image and its output image, the input resolution can be larger than the traditional training process, and also the computation cost.
To address this problem, we propose to reduce the computation cost by merging the early features of the input image and the output image.
Specifically, we feed the input image and the output image to the model in parallel, then add their features patch by patch after a few blocks, \eg, 3 blocks by default.
This design saves nearly half of the computation costs, but we find  no degradation in performance.

\paragraph{Loss Function}~~~
We use a simple regression loss to train \Ours.
Specifically, smooth-$\ell_1$~\cite{girshick2015fast} loss is computed on the masked pixels.
We also consider $\ell_1$ and $\ell_2$  in our experiments, and we find that the smooth-$\ell_1$ loss achieves the best performance.

\subsection{In-Context Inference}\label{sec:sub-inference}

At the first time, we design an in-context inference procedure, which is very flexible for performing both in-domain and out-of-domain vision tasks. As the input and output spaces of vision tasks have been unified as images, we can directly use the input/output paired images from the same task as the input condition (task prompts) to indicate which task to perform, and concatenate them with the input image and a masked image for completing the corresponding task.
Examples are shown in Figure~\ref{fig:teaser}.
This definition of task prompt does not require deep understanding of language instructions like previous approaches, but uses the visual signal as context which can be understood by the vision model and well matches the nature of the visual domain.

Also, different task prompts would lead to different results. Thus, how to select or generate a more suitable task prompt can be a new direction to explore. Here we present two simple baselines and we leave more explorations as the future work. The first baseline is to obtain a better prompt via selection, that we traverse the whole training set in a heuristic manner and select the best-performing example pair for each task. The second baseline is to generate a task prompt. We define the task prompt as the learnable tensors, freeze the whole model, and then use the training loss to optimize the task prompts.
We compare these two solutions with the random counterpart in the experiments with the visualizations, as shown in \S\ref{sec:exp-prompt}.

\begin{table*}[t]
    \centering
    \tablestyle{3.5pt}{1.15}
    \scalebox{0.88}{
    \begin{tabular}{r|cccccccccccc}
             & \multicolumn{3}{c}{ depth estimation} &  \multicolumn{1}{c}{semantic seg.} &  \multicolumn{1}{c}{panoptic seg.}  &  \multicolumn{1}{c}{keypoint det.} &
             \multicolumn{2}{c}{denoising} &
             \multicolumn{2}{c}{deraining} &
             \multicolumn{2}{c}{enhance.}\\
             & \multicolumn{3}{c}{ NYUv2} &  ADE-20K  & COCO  &  \multicolumn{1}{c}{COCO} & \multicolumn{2}{c}{SIDD} &
             \multicolumn{2}{c}{5 datasets} &
             \multicolumn{2}{c}{LoL} \\
              & RMSE $\downarrow$ & A.Rel $\downarrow$ & $\delta_1$ $\uparrow$ & mIoU $\uparrow$ & PQ $\uparrow$ & AP $\uparrow$  & PSNR $\uparrow$ & SSIM $\uparrow$ & PSNR $\uparrow$ & SSIM $\uparrow$  & PSNR $\uparrow$ & SSIM $\uparrow$ \\
            \shline
            \multicolumn{13}{c}{\small{\demphs{specialized models}}} \\
            \hline
            {\demphs{DenseDepth~\cite{alhashim2018high}}} & \demphs{0.465} & \demphs{0.123} & \demphs{0.846} & \demphs{-} & \demphs{-} & \demphs{-} & \demphs{-} & \demphs{-} & \demphs{-} & \demphs{-} & \demphs{-} & \demphs{-} \\
            {\demphs{BinsFormer~\cite{li2022binsdepth}}} & \demphs{0.330} & \demphs{0.094} & \demphs{0.925} & \demphs{-} & \demphs{-} & \demphs{-} & \demphs{-} & \demphs{-} & \demphs{-} & \demphs{-} & \demphs{-} & \demphs{-} \\
            {\demphs{UperNet-ViT-L~\cite{xiao2018upernet}}} & \demphs{-} & \demphs{-} &\demphs{-}  & \demphs{49.9} & \demphs{-} & \demphs{-} & \demphs{-} & \demphs{-} & \demphs{-} & \demphs{-} & \demphs{-} & \demphs{-} \\
            {\demphs{Mask2Former~\cite{mask2former}}} & \demphs{-} & \demphs{-} &\demphs{-}  & \demphs{57.7} & \demphs{57.8} & \demphs{-} & \demphs{-} & \demphs{-} & \demphs{-} & \demphs{-} & \demphs{-} & \demphs{-} \\
            {\demphs{DETR~\cite{carion2020detr}}} & \demphs{-} & \demphs{-} &\demphs{-}  & \demphs{-} & \demphs{45.6} & \demphs{-} & \demphs{-} & \demphs{-} & \demphs{-} & \demphs{-} & \demphs{-} & \demphs{-} \\
            {\demphs{HRNet~\cite{xiao2018sbaseline}}} & \demphs{-} & \demphs{-} & \demphs{-} & \demphs{-} & \demphs{-} & \demphs{76.3} & \demphs{-} & \demphs{-} & \demphs{-} & \demphs{-} & \demphs{-} & \demphs{-} \\
            {\demphs{HRFormer~\cite{mask2former}}} & \demphs{-} & \demphs{-} & \demphs{-} & \demphs{-} & \demphs{-} & \demphs{77.2} & \demphs{-} & \demphs{-} & \demphs{-} & \demphs{-} & \demphs{-} & \demphs{-} \\
            {\demphs{Uformer~\cite{uformer}}} & \demphs{-} & \demphs{-} & \demphs{-} & \demphs{-} & \demphs{-} & \demphs{-} & \demphs{39.89} & \demphs{0.960} & \demphs{-} & \demphs{-} & \demphs{-} & \demphs{-} 
            \\
            {\demphs{MPRNet~\cite{MPRNet}}} & \demphs{-} & \demphs{-} & \demphs{-} & \demphs{-} & \demphs{-} & \demphs{-} & \demphs{39.71} & \demphs{0.958} & \demphs{32.73} & \demphs{0.921} & \demphs{-} & \demphs{-} 
            \\
            {\demphs{MIRNet-v2~\cite{MIRNet}}} & \demphs{-} & \demphs{-} & \demphs{-} & \demphs{-} & \demphs{-} & \demphs{-} & \demphs{39.84} & \demphs{0.959} & \demphs{-} & \demphs{-} & \demphs{24.74} & \demphs{0.851}
            \\
            \shline
            \multicolumn{13}{c}{\small{{generalist framework, specialized models}}} \\
            \hline
            UViM~\cite{Kolesnikov2022UVIM} & 0.467 & - & - & - & 45.8  & - & - & - & - & - & - & -\\
            \shline
            \multicolumn{13}{c}{\small{{generalist models}}} \\
            \hline
            Unified-IO~\cite{Lu2022unifiedio} & 0.385 & - & - & - & - & - & - & - & - & - & - & -\\
            Pix2Seq v2~\cite{Chen2022pix2seq2} &-  & - & -  &  - & - & 64.8  & - & -  & - & - & - & -\\
            \Ours (ours) & 0.288 & 0.080 & 0.950 & 49.9 & 43.4 & 72.1 & 38.88 & 0.954 & 29.49 & 0.868 & 22.40 & 0.872 \\
        \end{tabular}
        }
        \caption{System-level comparison with the vision generalist models, and the recent best specialized models on seven representative tasks covering high-level visual understanding and low-level image processing. 
        We compare with the best results of each method. The backbones of the listed generalist methods are: ViT-L for UViM, Unified-IO$_{\rm XL}$ with 2925M parameters, ViT-B with another Transformer decoder for Pix2Seq v2, and ViT-L for \Ours.
}
        \label{tab:system}
\end{table*}

\begin{table*}[t]
    \centering
    \tablestyle{4.pt}{1.15}
        \scalebox{0.88}{
    \begin{tabular}{c|cccccccccccc}
             & \multicolumn{3}{c}{ depth estimation} &  \multicolumn{1}{c}{semantic seg.} &  \multicolumn{1}{c}{panoptic seg.}  &  \multicolumn{1}{c}{keypoint det.} &
             \multicolumn{2}{c}{denoising} &
             \multicolumn{2}{c}{deraining} &
             \multicolumn{2}{c}{enhance.}\\
             & \multicolumn{3}{c}{ NYUv2} &  ADE-20K  & COCO  &  \multicolumn{1}{c}{COCO} & \multicolumn{2}{c}{SIDD} &
             \multicolumn{2}{c}{5 datasets} &
             \multicolumn{2}{c}{LoL} \\
              & RMSE $\downarrow$ & A.Rel $\downarrow$ & $\delta_1$ $\uparrow$ & mIoU $\uparrow$ & PQ $\uparrow$ & AP $\uparrow$  & PSNR $\uparrow$ & SSIM $\uparrow$ & PSNR $\uparrow$ & SSIM $\uparrow$  & PSNR $\uparrow$ & SSIM $\uparrow$ \\
            \shline
            separate training & 0.327 & 0.090 & 0.930 & 47.2 & 41.3 & 72.5 & 38.43 & 0.956 & 28.28 & 0.844 & 21.67 & 0.850 \\
            joint training & 0.288 & 0.080 & 0.950 & 49.9 & 43.4 & 72.1 & 38.88 & 0.954 & 29.49 & 0.868 & 22.40 & 0.872 \\
        \end{tabular}
        }
        \caption{
        {Comparison of the models with the settings of  joint training {\it vs.}\  separate training on seven representative tasks.
        } 
        For the separate training setting, we train the model separately on each task with the same number of iterations as joint training on this task.}
        \label{tab:sing-task}
\end{table*}

\section{Experiments}\label{sec:experiments}

\subsection{Settings}

\paragraph{Datasets and input format}~~
NYUv2~\cite{nathan2012nyuv2} dataset consists of 464 indoor scenes captured by a Microsoft Kinect camera. The official training split (24K images) is used for training, and we report the Root Mean Square Error (RMSE), absolute mean relative error (A.Rel) and the percentage of inside pixels with different thresholds of $\delta$ on the 654 testing images from 215 indoor scenes.

\begin{table*}[t]
    \centering
    \tablestyle{4.pt}{1.15}
    \scalebox{0.88}{
    \begin{tabular}{c|cccccccccccc}
             & \multicolumn{3}{c}{ depth estimation} &  \multicolumn{1}{c}{semantic seg.} &  \multicolumn{1}{c}{panoptic seg.}  &  \multicolumn{1}{c}{keypoint det.} &
             \multicolumn{2}{c}{denoising} &
             \multicolumn{2}{c}{deraining} &
             \multicolumn{2}{c}{enhance.}\\
             & \multicolumn{3}{c}{ NYUv2} &  ADE-20K  & COCO  &  \multicolumn{1}{c}{COCO} & \multicolumn{2}{c}{SIDD} &
             \multicolumn{2}{c}{5 datasets} &
             \multicolumn{2}{c}{LoL} \\
              & RMSE $\downarrow$ & A.Rel $\downarrow$ & $\delta_1$ $\uparrow$ & mIoU $\uparrow$ & PQ $\uparrow$ & AP $\uparrow$  & PSNR $\uparrow$ & SSIM $\uparrow$ & PSNR $\uparrow$ & SSIM $\uparrow$  & PSNR $\uparrow$ & SSIM $\uparrow$ \\
            \shline
            random prompts & 0.291 & 0.080 & 0.949 & 49.3 & 43.1 & 71.8 & 38.46 & 0.953 & 29.26 & 0.865 & 22.31 & 0.871 \\
            searched prompts & 0.288 & 0.080 & 0.950 & 49.9 & 43.4 & 72.1 & 38.88 & 0.954 & 29.49 & 0.868 & 22.40 & 0.872  \\
            learned prompts & 0.286 & 0.080 & 0.949 & 49.9 & 43.3 & 72.2 & 38.71 & 0.954 & 29.56 & 0.870 & 22.38 & 0.872 \\
        \end{tabular}
        }
        \caption{Comparison of the models with random prompts, searched prompts, and learned prompts for different tasks.}
        \label{tab:prompt}
        \vspace{-1em}
\end{table*}

ADE20K~\cite{zhou2018ade} is a widely-used semantic segmentation dataset, covering a broad range of 150 semantic categories. It has 25K images in total, with 20K for training, 2K for validation, and another 3K for testing. We adopt the widely-used metric of mean IoU (mIoU) for evaluation.

MS-COCO~\cite{lin2014coco} contains approximately 118K training images and 5K validation images used for evaluation, with 80 ``things" and 53 ``stuff" categories. 
Panoptic segmentation task is evaluated on the union of ``things" and ``stuff" categories. During training, we generate the output images of semantic segmentation with 133 categories, and the class-agnostic instance segmentation with only ``things'' categories. During inference, we perform inference twice for each input image to obtain the results of semantic segmentation and class-agnostic instance segmentation, respectively, then merge them together to get the results of panoptic segmentation. We report panoptic quality as the measure.

For human keypoint detection, we use the standard person detection results from Simple Baseline~\cite{xiao2018sbaseline}, which use the standard splits on COCO with 15K training samples and validation set for evaluation, and report the AP based on OKS as the evaluation metric~\cite{xiao2018sbaseline}.

Image restoration tasks are evaluated on several popular benchmarks, including SIDD~\cite{sidd} for image denoising, LoL~\cite{lol} for low-light image enhancement, and the merged deraining dataset~\cite{MIRNet} for deraining. More details about these datasets are provided in Table~\ref{table:lowlevel_datasets}.

\paragraph{Training details}~~
During training, we employ an AdamW optimizer~\cite{kingma2014adam} with a \emph{cosine} learning rate scheduler, and train for 54K iterations. The training hyper-parameters are: the batch size as 2048, base learning rate as 1{$e$}$-3$, weight decay as 0.05, $\beta_1=0.9$, $\beta_2=0.999$, drop path~\cite{huang2016deep} ratio as 0.1, warm-up for 5.4K iterations. 
A light data augmentation strategy is used: random resize cropping with scale range of $ [0.3, 1] $
and a aspect ratio range of $ [\nicefrac{3}{4}, \nicefrac{4}{3}]$, followed by a random flipping.
The input image size is $448\times 448$ by default. 
The sampling weight for each task is $0.1$ (NYUv2 depth estimation), $0.2$ (ADE-20K semantic segmentation), $0.15$ (COCO class-agnostic instance segmentation), $0.25$ (COCO semantic segmentation), $0.2$ (COCO human keypoint detection), $0.15$ (image denoising), $0.05$ image deraining, and $0.05$ (low-light image enhancement).

\subsection{Results}\label{sec:exp-main}

\paragraph{System-level comparison}~~
With the corresponding task prompts, we compare our approach, \Ours, with recent best vision generalist models and specialized models on seven representative tasks, shown in Table~\ref{tab:system}. Without task-specific design, our \Ours sets new records on NYUv2 depth estimation, outperforming not only vision generalists, such as Unified-IO~\cite{Lu2022unifiedio} and UViM~\cite{Kolesnikov2022UVIM}, but also the state-of-the-art specialized model, \eg, BinsFormer~\cite{li2022binsdepth}. 
For COCO keypoint detection, \Ours significantly outperforms the generalist model Pix2Seq v2~\cite{Chen2022pix2seq2} by 7.3 AP.
For ADE-20K semantic segmentation, COCO panoptic segmentation, and the three low-level image processing tasks, 
our model achieves comparable performance to those well-designed task-specific models. 
There is still much room for boosting our approach compared to other well-designed specialized models. 
For example, our default input image size is $448\times 448$ while those specialized panoptic segmentation models use a much larger resolution, \eg, $1024\times 1024$ in Mask2Former~\cite{mask2former}.
Achieving state-of-the-art performance on every task is not the major goal of this paper, and we leave this as future work.

\begin{table}
\small
  \centering
\resizebox{0.51\linewidth}{!}{
  \begin{tabular}{l|c}
    \toprule
    Model &  mIoU~$\uparrow$  \\
    \midrule
    MAE-VQGAN~\cite{Bar2022VisualPrompt}  & 58.3 \\ %
    Painter & 62.3 \\
    \bottomrule
  \end{tabular}
  }
    \caption{Quantitative results on open-vocabulary FSS-1000.}
    \label{tab:fss}
    \vspace{-0.5em}
\end{table}

\paragraph{Joint training vs.\  separate training}~~
We compare two training settings, \ie, joint training and separate training, in Table~\ref{tab:sing-task}. For the separate training setting, we train each model separately on each task with the same number of iterations and architectures as joint training on this task. 
We can observe that models with joint training generally outperform that with separate training on most of the tasks, indicating that our in-context training approach with the unification of output spaces can somewhat benefit them from each other.
But conflict may still exist, \eg, joint training performs slightly worse on keypoint detection.
Exploring the relationships between tasks  in our simple and unified framework could be an interesting and promising direction.

\paragraph{Qualitative results}~~
To demonstrate the capability of our generalist model in an intuitive perspective, we visualize the task output of the selected images from the validation set of several in-domain tasks, such as semantic segmentation, depth estimation, instance segmentation, human keypoint detection, image denoising, image deraining, and low-light image enhancement. 
As shown in Figure~\ref{fig:vis}, \Ours can make very accurate predictions on all these tasks.

\subsection{Prompt Tuning}\label{sec:exp-prompt}

Different task prompts would lead to varied results. Here we compare three simple baselines, `random', `searched' and `learned' prompts. Random prompts denote to randomly select one example from that task as the prompt.
Searched prompt denotes to traverse the  training set in a heuristic manner and select the best-performing example pair for each task. Learned prompt denotes that we define the task prompt as the learnable tensors, freeze the whole model, and then use the training loss to optimize the task prompts.

\begin{figure}
    \centering
    \includegraphics[width=0.9\linewidth]{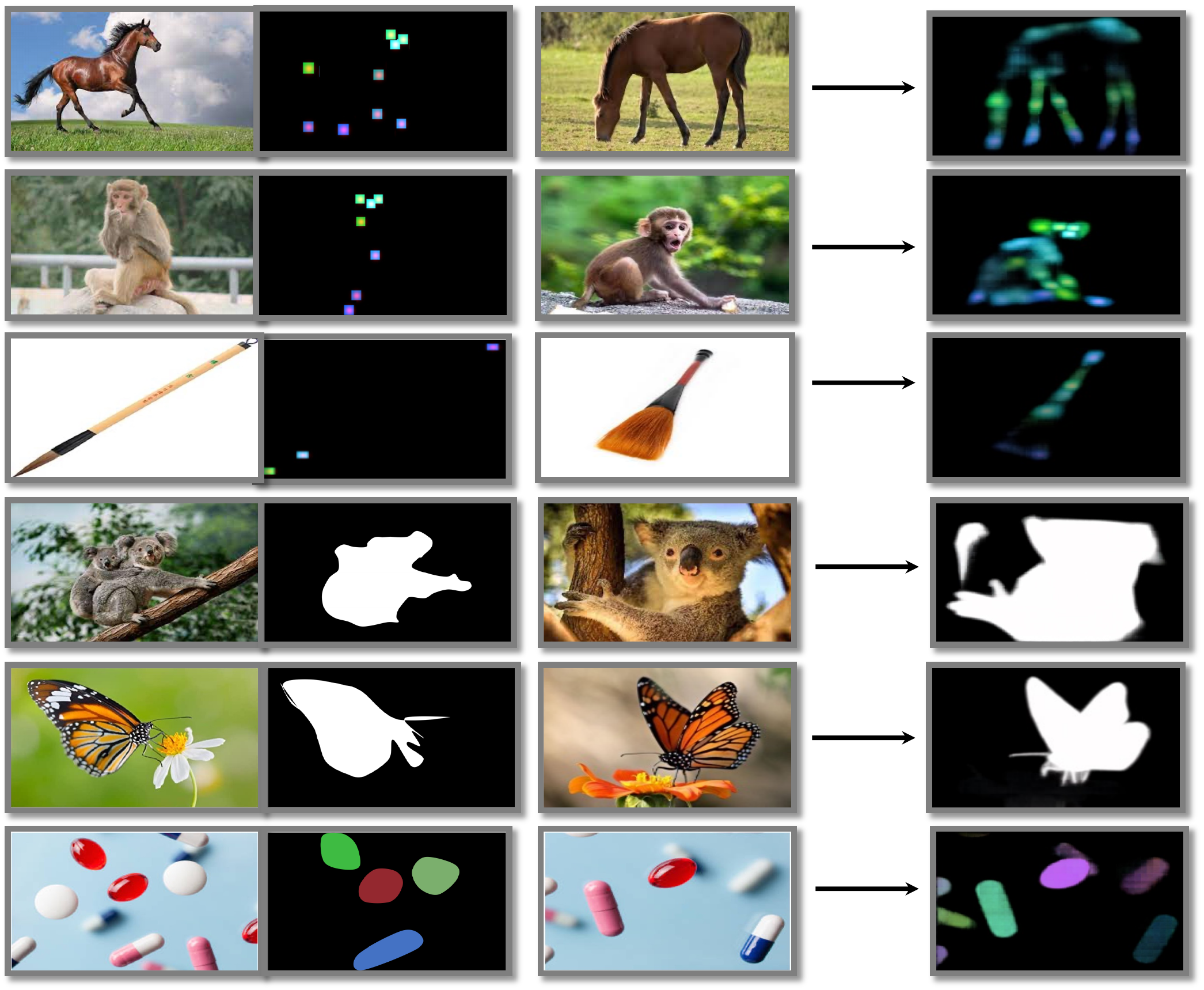}
    \vspace{-0.3em}
    \caption{Visualizations of examples and predictions obtained by \Ours, such as open-vocabulary keypoint detection (e.g., horse, monkey and broom), object segmentation (koala and butterfly) and instance segmentation (tablets).}
    \vspace{-2pt}
    \label{fig:emergent_vis}
\end{figure}

The qualitative comparisons are shown in Table~\ref{tab:prompt} . We can see that the model using the searched and learned prompts perform better than that with random ones, which indicates that the optimization process helps the task to find a more suitable prompt.  
Also, the model with random prompts works relatively well, indicating the robustness of our model.
In Figure~\ref{fig:selectedprompt}, we show the visualizations of the learned prompt images for different tasks.

It is very promising that the performance of a task can be improved by optimizing the input prompt. This provides the possibility that if a category or task does not appear in the training data, we can still achieve relatively good performance on this task by optimizing the prompts, without tuning the parameters of the model, which is also one core advantage of in-context learning.

\subsection{Generalization}\label{sec:exp-properties}

The core property of in-context learning is that it allows the model to rapidly adapt to various tasks with only a handful of prompts and examples. %
Here, we explore this capability via visualizations, shown in Figure~\ref{fig:teaser},  Figure~\ref{fig:emergent_vis}, and Figure~\ref{fig:supp_emergent_vis}.
From these visualizations, we find that our model can perform the task seen in training but with input images that their categories are unseen in training, such as open-vocabulary keypoint detection (\eg, monkey, horse and broom), object segmentation (koala) and instance segmentation (tablets).
In addition, we quantitatively evaluate \Ours on a few-shot segmentation benchmark which requires to segment objects of 1k novel classes.
\Ours largely outperforms a concurrent work ~\cite{Bar2022VisualPrompt}, as reported in Table~\ref{tab:fss},

\section{Discussion and Conclusion}

In this work, for the first time, we explore how to perform in-context visual learning and present a vision-centric solution in which the context is defined as visual signals. Thus, the models are granted with the capability to perform tasks conditioned on the paired data from that task, that is, in-context inference. 
Our approach achieves highly competitive performance on a representative and diverse set of seven tasks.

This work is not without drawbacks. 
First, there is still %
much room for boosting our approach especially on the difficult task of panoptic segmentation, comparing to the specialized models. In addition, since our approach is designed based on visual signals as contexts, this general interface does not seem natural for modeling language signals. How to model discrete language signals as continuous ones seems to be an impressive direction, and some work has started to emerge recently.

While there are previous approaches that hope to use general interfaces to solve multiple vision tasks, we may be 
the first to grant models the ability to learn and complete tasks in context, which does have the opportunity to handle out-of-domain tasks. We hope this work will draw attention to this promising direction, and we believe that the best GPT-3 moment in the vision field is yet to come.

\section*{Acknowledgement}
\label{sec:ack}
This project is supported by the National Key R\&D Program of China (2020AAA0105200).
We would like to thank Hanxiao Qu, Yemin Shi, Yan Tian, and Xigang Cao for their help on the GPU resources, Xinxin Liu and Kai Lu for beautifying the figures, as well as other colleagues at Beijing Academy of Artificial Intelligence for support throughout this project.

{\small
\bibliographystyle{ieee_fullname}
\bibliography{Painter}
}

\appendix

\section*{Appendix}

\renewcommand{\thefigure}{S\arabic{figure}}
\setcounter{figure}{0}
\renewcommand{\thetable}{S\arabic{table}}
\setcounter{table}{0}

\section{Additional Implementation Details}
In this section, we provide additional details of the data preparation and pose-processing for different tasks. PyTorch-style pseudo-code is provided to better illustrate the implementation details.
They can be done with a few operations in implementation.
For each task, Painter doesn't involve more complex post-processing compared to the specialist methods.

\paragraph{Semantic segmentation}
As described in Section 3.1 of the main paper, we formulate different semantic categories using different colors in RGB space. To this end, we define the background and ignore areas as black, \ie, pixels in color (0, 0, 0), and generate the colors for foreground categories using the pseudo-code elaborated in Figure~\ref{fig:gen_color}. 

During inference, to decode the output image to a single-channel ID map where each pixel represents a class ID, we compute the L1 distance between each output pixel and the pre-defined colors for each semantic category, and take the ID of the closest color as the predicted category. The pseudo-code for the post-processing is illustrated in Figure~\ref{fig:post_segm}.

\lstset{
  backgroundcolor=\color{white},
  basicstyle=\fontsize{7.5pt}{8.5pt}\fontfamily{lmtt}\selectfont,
  columns=fullflexible,
  breaklines=true,
  captionpos=b,
  commentstyle=\fontsize{8pt}{9pt}\color{codegray},
  keywordstyle=\fontsize{8pt}{9pt}\color{codegreen},
  stringstyle=\fontsize{8pt}{9pt}\color{codeblue},
  frame=tb,
  otherkeywords = {self},
}
\begin{figure}[ht]

\begin{lstlisting}[language=python]
def generate_colors_dict(b, K):
    # b: the number of used values in a single channel
    # K: the number of classes
    
    m = 255 // b  # get margin
    colors = []
    for class_id in range(K):
        # compute margin multiplier
        r_mult = class_id // b**2
        g_mult = (class_id %
        b_mult = class_id %
        # compute r, g, b values
        r = 255 - r_mult * m
        g = 255 - g_mult * m
        b = 255 - b_mult * m
        colors.append((r, g, b))

    return colors

\end{lstlisting}
\caption{Pseudo-code for generating colors.
\texttt{$/\!/$}: floor division;
\texttt{$\%$}: mod operation.
}
\label{fig:gen_color}
\end{figure}

\lstset{
  backgroundcolor=\color{white},
  basicstyle=\fontsize{7.5pt}{8.5pt}\fontfamily{lmtt}\selectfont,
  columns=fullflexible,
  breaklines=true,
  captionpos=b,
  commentstyle=\fontsize{8pt}{9pt}\color{codegray},
  keywordstyle=\fontsize{8pt}{9pt}\color{codegreen},
  stringstyle=\fontsize{8pt}{9pt}\color{codeblue},
  frame=tb,
  otherkeywords = {self},
}
\begin{figure}[ht]
\begin{lstlisting}[language=python]
def forward(image, colors):
    # image: (H, W, 3)
    # colors: (K, 3), where K is the number of classes

    # get distance between pixels and pre-defined colors
    dist = (image.view(H, W, 1, 3) - colors.view(1, 1, K, 3)).abs().sum(-1)  # (H, W, K)
    segm = dist.argmin(dim=-1)  # (H, W)
    
    return segm
    
\end{lstlisting}
\caption{Pseudo-code of semantic segmentation post-processing.
}
\label{fig:post_segm}
\end{figure}

\paragraph{Keypoint detection} 
For keypoint detection, the output image consists of the R channel which denotes the class-agnostic heatmaps and the G/B channels that represent the keypoint categories. %
As illustrated in Figure~\ref{fig:post_pose}, we convert the output image to a 17-channel heatmap, and follow the commonly used post-processing~\cite{xiao2018sbaseline, sun2019hrnet} to obtain the final keypoint locations.

\lstset{
  backgroundcolor=\color{white},
  basicstyle=\fontsize{7.5pt}{8.5pt}\fontfamily{lmtt}\selectfont,
  columns=fullflexible,
  breaklines=true,
  captionpos=b,
  commentstyle=\fontsize{8pt}{9pt}\color{codegray},
  keywordstyle=\fontsize{8pt}{9pt}\color{codegreen},
  stringstyle=\fontsize{8pt}{9pt}\color{codeblue},
  frame=tb,
  otherkeywords = {self},
}
\begin{figure}[ht]
\begin{lstlisting}[language=python]
def forward(image, colors):
    # image: (H, W, 3)
    # colors: (K, 3), where K is the number of keypoints

    # r for heatmaps and gb for keypoint classes 
    r = images[..., 0]  # (H, W)
    gb = images[..., 1:]  # (H, W, 2)
    
    # get keypoint class of each pixel
    dist = (gb.view(H, W, 1, 2) - colors.view(1, 1, K, 2)).abs().sum(-1)  # (H, W, K)
    segm = dist.argmin(dim=-1)  # (H, W)

    for idx in range(K):
        mask = segm == idx
        heatmap = mask * r  # (H, W)
        heatmaps.append(heatmap)
    heatmaps = stack(heatmaps)  # (K, H, W)

    return heatmaps
    
\end{lstlisting}
\caption{Pseudo-code of keypoint detection post-processing.
\texttt{stack}: concatenates a sequence of tensors along a new dimension.
}
\label{fig:post_pose}
\end{figure}

\begin{table*}[!t]
\begin{center}
\setlength{\tabcolsep}{0.8pt}
\scalebox{0.8270}{
\begin{tabular}{l | c c c c c c c | c | c }
 Tasks & \multicolumn{7}{c|}{Deraining} & \multicolumn{1}{c|}{Enhance.} & \multicolumn{1}{c}{Denoising}\\
\toprule[0.15em]
Datasets  &   Rain14000~\cite{fu2017removing}  &  Rain1800~\cite{yang2017deep}  & Rain800~\cite{zhang2019image} & Rain100H~\cite{yang2017deep} & Rain100L~\cite{yang2017deep} & Rain1200~\cite{zhang2018density} & Rain12~\cite{li2016rain} & LoL~\cite{lol}  & SIDD~\cite{sidd} \\
Train Samples     &   11200      &  1800      & 700     & 0        & 0        & 0        & 12     & 485 & 320   \\
Test Samples      &   2800       &  0         & 100     & 100      & 100      & 1200     & 0      & 15 & 40 \\
\midrule
Testset Rename & Test2800 & - & Test100 &	Rain100H	& Rain100L & Test1200  & - & - & - \\
\end{tabular}}
\caption{\small Dataset description for image restoration tasks.}
\label{table:lowlevel_datasets}
\end{center}
\end{table*}

\begin{table*}[t] 
\centering
\subfloat[
\textbf{Patch merging}
\label{tab:ablation-merge}
]{
    \centering
    \tablestyle{4pt}{1.15}
   \begin{tabular}{c|cc}
            merging?
              & mIoU & mAcc \\
            \shline
            \xmark &   39.7 & 51.1 \\
            \cmark &  41.2 & 53.0 \\
        \end{tabular}
}
\subfloat[
\textbf{Encoder}
\label{tab:ablation-encoder}
]{
    \centering
    \tablestyle{4pt}{1.15}
   \begin{tabular}{c|cc}
            backbone
             &  mIoU & mAcc \\
            \shline
            ViT-B & 31.4 & 41.4 \\
            ViT-L & 41.2 & 53.0 \\
    \end{tabular}
}
\vspace{1em}
\subfloat[
\textbf{Head type}
\label{tab:ablation-decoder}
]{
    \centering
    \tablestyle{4pt}{1.15}
   \begin{tabular}{c|cc}
            head 
              & mIoU & mAcc \\
            \shline
            linear & 38.6 & 50.2 \\
            light  & 41.2 & 53.0 \\
        \end{tabular}
}
\subfloat[
\textbf{Loss function}
\label{tab:ablation-loss}
]{
    \centering
    \tablestyle{4pt}{1.15}
   \begin{tabular}{c|cc}
            loss
             &  mIoU & mAcc \\
            \shline
            $\ell_1$ & 40.6& 52.5 \\
            $\ell_2$ &  26.3 & 37.7 \\
            smooth-$\ell_1$ & 41.2 & 53.0 \\
        \end{tabular}
}
        \caption{Ablation study on ADE-20K semantic segmentation. (a)  merging patch after three transformer blocks; (b) encoder; (c) head type; (d) loss function.
        }
        \label{tab:ablation}
\end{table*}

\paragraph{Panoptic segmentation}
As described in Section 3.2, we decompose the panoptic segmentation task into semantic segmentation and class-agnostic instance segmentation.
During training, the semantic segmentation sub-task uses the same setting as the semantic segmentation on ADE-20K~\cite{zhou2018ade}, except that we set the base $b=7$ when assigning colors.
Similar to the color generation process of semantic segmentation, we generate colors for each location category~\cite{wang2020solo} used in class-agnostic instance segmentation. The color of each instance mask is determined by the location of its center.

During inference, the semantic ID map can be obtained using the post-processing described in Figure~\ref{fig:post_segm}, while the class-agnostic instance masks are generated by thresholding the distance between predicted colors and the pre-defined colors for location categories. Matrix NMS~\cite{wang2020solov2} is adopted to remove duplicate instance predictions.
We apply the majority vote of pixels from the semantic prediction to get the semantic class for each instance mask, as illustrated in Figure~\ref{fig:post_inst_vote}.
Finally, we follow Panoptic FPN~\cite{panopticfpn} to merge the semantic segmentation and the instance segmentation predictions to obtain the panoptic segmentation results.

\lstset{
  backgroundcolor=\color{white},
  basicstyle=\fontsize{7.5pt}{8.5pt}\fontfamily{lmtt}\selectfont,
  columns=fullflexible,
  breaklines=true,
  captionpos=b,
  commentstyle=\fontsize{8pt}{9pt}\color{codegray},
  keywordstyle=\fontsize{8pt}{9pt}\color{codegreen},
  stringstyle=\fontsize{8pt}{9pt}\color{codeblue},
  frame=tb,
  otherkeywords = {self},
}
\begin{figure}[ht]
\begin{lstlisting}[language=python]
def forward(segm_dist, inst_masks):
    # segm_dist: (H, W, K), where K is the number of thing classes
    # inst_masks: (N, H, W), where N is the number of instances
    
    # turn distances to scores
    segm_scores =  1. - semseg_dist / max(semseg_dist)
    # majority vote
    class_probs = einsum("nhw,hwk->nk", inst_masks, segm_scores)  # (N, K)
    pred_classes = class_probs.argmax(dim=-1)  # (N,)
    
    return pred_classes

\end{lstlisting}
\caption{Pseudo-code of majority voting for labeling each instance mask.
\texttt{einsum}: Einstein summation.
}
\label{fig:post_inst_vote}
\end{figure}

\begin{figure}[ht]
    \centering
    \includegraphics[width=1.\linewidth]{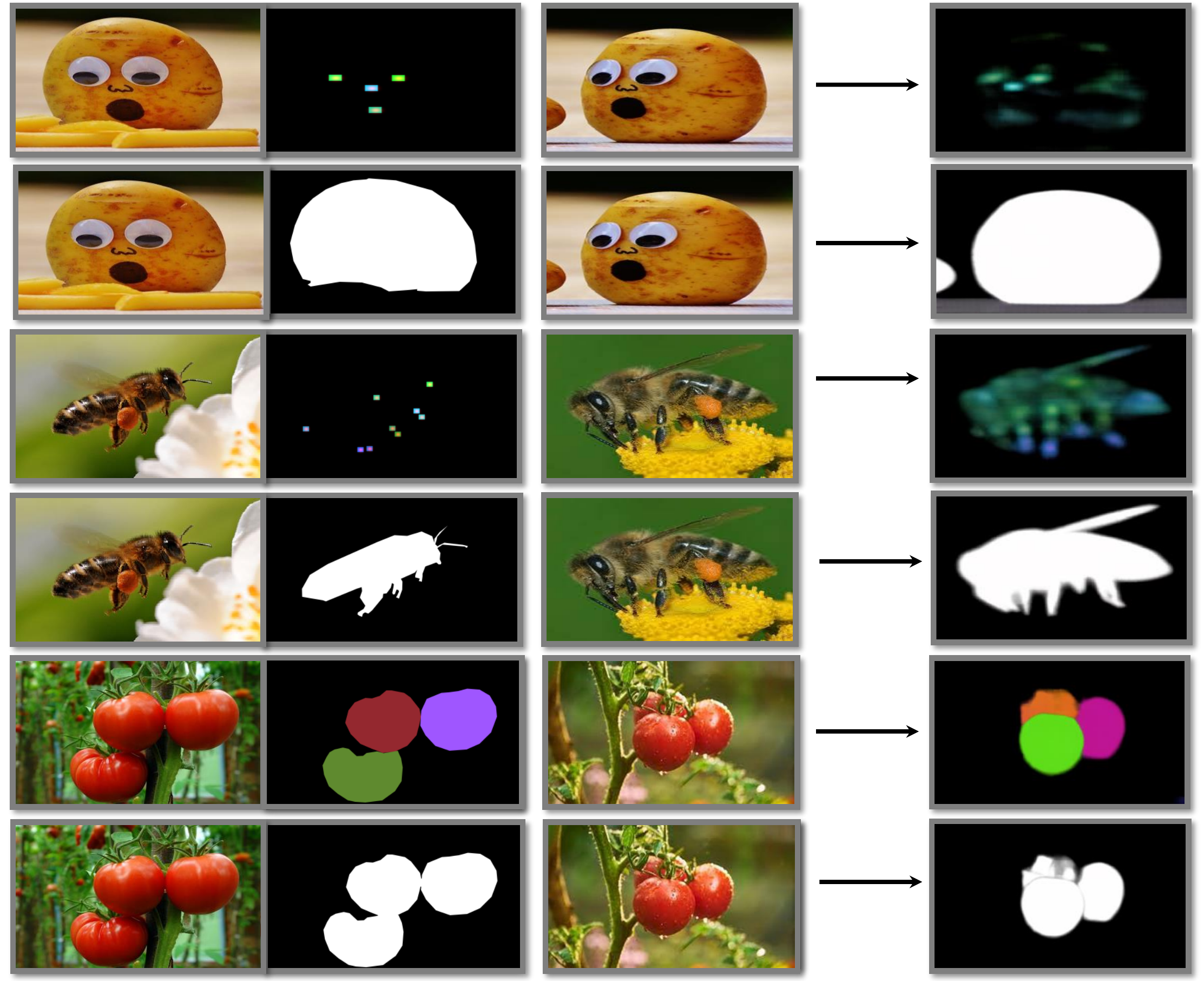}
    \vspace{-1em}
    \caption{More visualizations. The visualized tasks include keypoint detection, object segmentation and instance segmentation on potato, bee, and tomato.
    }
    \vspace{-1em}
    \label{fig:supp_emergent_vis}
\end{figure}

\paragraph{Image restoration} The detailed statistics of the datasets that are used for image restoration are shown in Table~\ref{table:lowlevel_datasets}.

\begin{figure*}[ht]
    \centering
    \includegraphics[width=.95\linewidth]{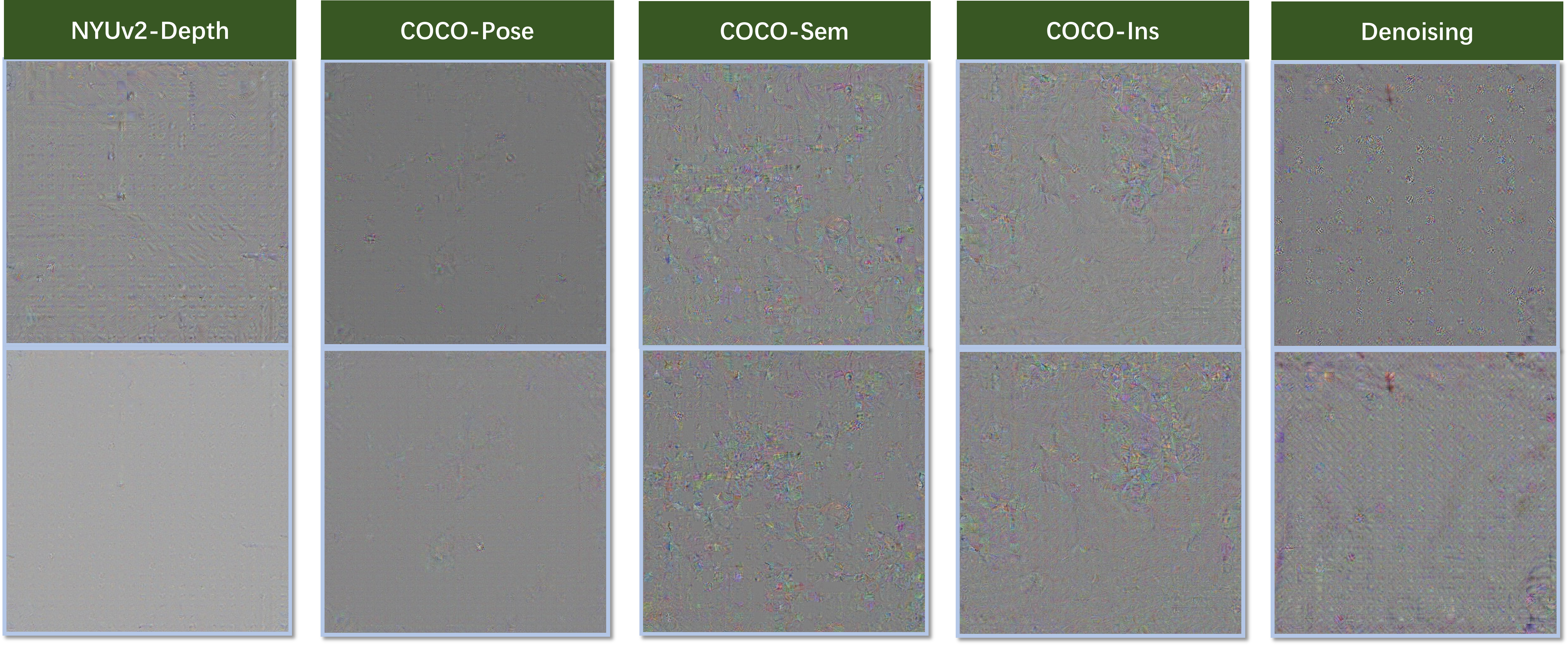}
    \caption{
    Visualizations of the learned prompts for different tasks. Each column denotes a prompt pair for a task. The first and second rows are input and output images respectively. For visualization, each prompt is normalized to an RGB image with values between 0 and 255. Different tasks show different patterns. Users can take the prompt images and feed them to the \Ours model to enable the corresponding application.}
    \label{fig:selectedprompt}
\end{figure*}

\section{Additional Results}

We report the results of ablation experiments on several components of our framework, with a shorter schedule of 3k iterations and other hyper-parameters unchanged on semantic segmentation of ADE-20K.

\paragraph{Merging patches} During training, each input sample consists of both the input image and output image, which results in high memory cost and significantly slows down the training process. 
We reduce nearly half of the computation costs by merging the early features of the input image and the output image, \ie, adding their features patch by patch after a three blocks.
Table~\ref{tab:ablation-merge} shows that this new design even incurs performance increase. 
We argue that this design further provides the pixel-to-pixel correspondence between the input and its output via stacking them together. But in the original setting, these relationships need to be learned by the model, which will make the optimization more difficult especially in a short schedule.

\paragraph{Encoder} We adopt standard Vision Transformer (ViT)~\cite{dosovitskiy2020vit} with different model sizes as the encoder, including ViT-base and ViT-large. Results are shown in Table~\ref{tab:ablation-encoder}. We find that the model with ViT-L outperforms that with ViT-B by very large margins.
This observation is intuitive, that generally larger models yield better performance. For generalist models, they can use more data but with less task-specific prior on method design, thus may require more model capacity than task-specific models.

\paragraph{Head}
We use a light three-layer head that consists of a linear (1$\times$1 convolution) layer, a 3$\times$3 convolution layer, and another linear layer, to map the feature of each patch to its original resolution, \eg, $16\times 16\times 3$. 
The feature of each patch is the concatenation of the  4 feature maps evenly sampled from the transformer blocks.
As shown in Table~\ref{tab:ablation-decoder}, the light head achieves clear gains over the baseline with only a linear layer.

\paragraph{Loss function}
\Ours uses a simple pixel regression loss to learn all the tasks.
In Table~\ref{tab:ablation-loss}, we compare different regression loss functions, including $\ell_1$, $\ell_2$, and smooth-$\ell_1$.
We adopt smooth-$\ell_1$ by default as it achieves the best performance and is also more stable during training.

\section{Additional Visualization}

In this section, we provide more visualizations.
As shown in Figure~\ref{fig:supp_emergent_vis}, \Ours performs in-context inference according to different prompt images. Note that \Ours is never trained to solve these tasks during training, \eg, keypoint detection of potato, object segmentation of bee, and instance segmentation of tomato.

\end{document}